\documentclass{article}

\PassOptionsToPackage{numbers, compress}{natbib}

\usepackage[preprint]{main}

\usepackage[utf8]{inputenc} 
\usepackage[T1]{fontenc}    
\usepackage{hyperref}       
\usepackage{url}            
\usepackage{booktabs}       
\usepackage{amsfonts}       
\usepackage{nicefrac}       
\usepackage{microtype}      
\usepackage{graphicx}
\usepackage{algorithm}
\usepackage{amsmath}
\usepackage{tabularx}
\usepackage{subcaption}
\usepackage{colortbl}
\usepackage[table]{xcolor}
\newfloat{figtab}{htb}{fgtb}
\makeatletter
  \newcommand\figcaption{\def\@captype{figure}\caption}
  \newcommand\tabcaption{\def\@captype{table}\caption}
\makeatother
\usepackage{multirow}
\usepackage{wrapfig}
\usepackage{amssymb}
\usepackage{pifont}
\usepackage{threeparttable}

\usepackage{xcolor}         
 \usepackage{amsmath}
\usepackage{colortbl}
\definecolor{mygray}{rgb}{0.941, 0.941, 0.956}

\def\logo{\makebox[30pt][l]{\raisebox{-0.6ex}{\includegraphics[height=26pt]{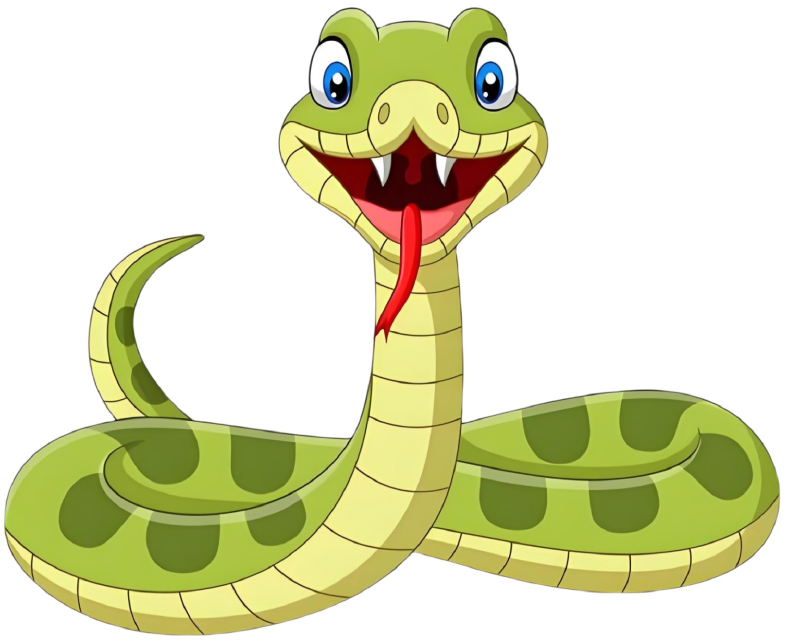}}}\hspace{5pt}}

\title{\logo  ML-Mamba: Efficient Multi-Modal Large Language Model Utilizing Mamba-2}

\author{%
Wenjun Huang$^{1,*}$ \quad Jiakai Pan$^{1,*}$ \quad Jiahao Tang$^{1}$ \quad Yanyu Ding$^{2}$ \\[0pt]
\textbf{Yifei Xing$^{3}$ \quad Yuhe Wang$^{1}$ \quad Zhengzhuo Wang$^{1}$ \quad Jianguo Hu$^{1,\dag}$}\\[0pt]
$^1$Sun Yat-sen University, $^2$Dongguan University of Technology \\[0pt] 
$^3$University of the Chinese Academy of Sciences \\[0pt] 
{\tt\small  huangwj98@mail2.sysu.edu.cn \quad hujguo@mail.sysu.edu.cn} \\[0pt]
Project URL:  
{\small \url{https://wenjunhuang94.github.io/ML-Mamba}} \\[0pt]
$^*$These authors contributed equally to this work. \\[0pt]
$^\dag$Corresponding author: Jianguo Hu
}

\usepackage{tcolorbox}
\begin{document}

\maketitle

\begin{abstract}
Multimodal Large Language Models (MLLMs) have attracted much attention for their multifunctionality. However, traditional Transformer architectures incur significant overhead due to their secondary computational complexity. To address this issue, we introduce ML-Mamba, a multimodal language model, which utilizes the latest and efficient Mamba-2 model for inference. Mamba-2 is known for its linear scalability and fast processing of long sequences. We replace the Transformer-based backbone with a pre-trained Mamba-2 model and explore methods for integrating 2D visual selective scanning mechanisms into multimodal learning while also trying various visual encoders and Mamba-2 model variants. Our extensive experiments in various multimodal benchmark tests demonstrate the competitive performance of ML-Mamba and highlight the potential of state space models in multimodal tasks. The experimental results show that: (1) we empirically explore how to effectively apply the 2D vision selective scan mechanism for multimodal learning. We propose a novel multimodal connector called the Mamba-2 Scan Connector (MSC), which enhances representational capabilities. (2)  ML-Mamba achieves performance comparable to state-of-the-art methods such as TinyLaVA and MobileVLM v2 through its linear sequential modeling while faster inference speed; (3) Compared to multimodal models utilizing Mamba-1, the Mamba-2-based ML-Mamba exhibits superior inference performance and effectiveness.
\end{abstract}

\section{Introduction}
\label{sec:intro}
The emergence of Large Language Models (LLMs) has profoundly changed the landscape of natural language understanding tasks. Unlike early methods that relied on medium-sized task specific models, recent advances have shifted towards using general large-scale models, especially after the success of systems such as ChatGPT. It has been proven that expanding the scale of language models and increasing data volume can bring many advantages, including enhancing the performance of different tasks and improving the sample efficiency of out of distribution generalization~\cite{he2024multi}.

However, traditional LLMs are limited to interacting through language, which limits their adaptability to handling more diverse tasks. Multi modal understanding that integrates visual and textual information is crucial for improving the ability of models to effectively respond to real-world challenges. Therefore, researchers are actively expanding large-scale language models to integrate multimodal information processing capabilities. Visual language models (VLMs) such as GPT-4~\cite{openai2024gpt4}, LLaMA adapter~\cite{gao2023llamaadapter}, and LLaVA~\cite{liu2023llava, liu2023llava1.5} have been developed to enhance LLM's visual comprehension ability. These VLMs are fundamental models for handling a range of tasks, including visual question answering (VQA), image captioning, and visual content generation.

Despite achieving success, previous research has mainly focused on reducing the parameters of language models while preserving the Transformer architecture. However, this method does not solve the inherent problem of low computational efficiency in Transformer's self attention mechanism, which is quadratic with sequence length. To address this bottleneck, the latest research work has designed a new architecture (Mamba-2), whose core layer is an improvement of Mamba selective SSM. The state space model (SSM) has been widely studied as an effective alternative solution. SSM combines elements of Recurrent Neural Networks (RNNs) and Convolutional Neural Networks (CNNs), providing linear scaling of sequence length and effective training and inference. It is 2-8 times faster and continues to compete with Transformers in language modeling.

To this end, this article proposes a new perspective, directly using the state space model (SSM) as the backbone. Specifically, we use the Mamba-2 language model as the basic model of our VLM. In this article, we introduce ML-Mamba, a work that applies state space models to multimodal learning tasks. Our method utilizes a pre-trained Mamba-2 language model as the backbone, replacing traditional Transformer-based models such as LLaMA~\cite{touvron2023llama}. We further enhanced ML-Mamba through a novel multimodal connector called Mamba-2 Scan Connector (MSC) architecture, which includes a Mamba-2 visual selective scanning module (MVSS) and a SwiGLU module specifically both designed for 2D causal modeling of enriched visual sequences. The MVSS module explores two different scanning mechanisms: bidirectional scanning mechanism (BSM) and cross scanning mechanism (CSM). In addition, we investigated the combination of different visual encoders, variants of pre-trained Mamba-2 language models, and multimodal connectors to optimize the integration of visual and linguistic information.

Extensive experiments conducted on a range of multimodal learning benchmarks demonstrate the efficacy of ML-Mamba. Our model not only achieves competitive performance with other similarly sized small multimodal large-scale language models (MLLMs) but also surpasses larger MLLMs on several prominent benchmark tests, including LLaVA v1.5~\cite{liu2023llava1.5} versions 7B and 13b.

The major contributions of this paper are three-fold:

\begin{itemize}
\item We propose a novel and efficient method, i.e., ML-Mamba, which explores and utilizes multimodal learning tasks combined with the latest Mamba-2. Compared to the multimodal model adopting the original Mamba, the multimodal large-scale language model based on Mamba-2 has higher inference performance and effectiveness. Meanwhile, ML-Mamba also provides a new framework choice for multimodal large-scale language models beyond Transformer-based architectures.
\item We empirically explore the impact of different components in ML-Mamba and propose a novel multimode connector called Mamba-2 Scan Connector (MSC). MSC includes the Mamba-2 Visual Selective Scanning (MVSS) module and the SwiGLU module, which enhance representational capabilities.
\item We conduct extensive experiments on different multimodal learning benchmarks.
The numerical results show that ML-Mamba achieves competitive performance compared to existing multimodal large-scale language models.
\end{itemize}

\section{Related Work}
\label{sec:related work}

\subsection{Large Language Models (LLMs)}
In recent years, significant breakthroughs have been made in natural language processing tasks~\cite{ jia2024gpt4mts, kim2024structure}, characterized by large model scales, typically containing billions of parameters, and training using massive datasets. GLM~\cite{du2022glm}, LLaMA~\cite{touvron2023llama}, Alpaca~\cite{taori2023alpaca}, Vicuna~\cite{vicuna2023} and other instruction fine-tuning versions have emerged one after another, with the goal of being comparable to the proprietary InstructGPT model without public access. At the same time, due to the significant computational requirements of large language models, research trends have shifted towards exploring the possibility of smaller scale models, such as Stable LM~\cite{bellagente2024stable}, TinyLaMA ~\cite{zhang2024tinyllama}, and Phi~\cite{gunasekar2023textbooks, textbooks2}, which have parameter sizes below 3 billion but can achieve comparable results to large models through high-quality data and feasible training methods.

\subsection{State Space Models (SSMs)}
State Space Models (SSMs) have demonstrated excellent performance in areas such as long sequence modeling, image generation, and reinforcement learning. A notable feature of SSMs is their ability to perform efficient autoregressive inference like Recurrent Neural Networks (RNNs) while also being able to process entire input sequences in parallel like attention-based Transformers, thus enabling efficient training. Despite their efficiency, SSMs achieve good results in various sequence modeling tasks. Specifically, Albert et al.~\cite{ssm1} proposed a structured state space sequence model for time series analysis. Goel et al.~\cite{goel2022sashimi} applied SSMs to audio generation and achieved satisfactory performance. Additionally, the H3 model~\cite{fu2023hungry} was introduced to bridge the gap between SSMs and Transformers in language modeling.

In recent months, a new selective state space model called Mamba~\cite{gu2023mamba} has been proposed as a strong competitor to the Transformer architecture. Compared to LLMs of the same capacity, language models based on Mamba have shown competitive performance, faster inference speeds, and the ability to scale linearly over time with constant memory usage. In May 2024, the latest Mamba architecture (Mamba-2)~\cite{Dao2024TransformersAS} was introduced, featuring an improved core layer of the Mamba selective SSM, which is 2-8 times faster while continuing to compete with Transformers in language modeling.

\subsection{Multimodal Large Language Model (MLLM)}
The Multi Modal Large Language Model (MLLM) combines visual and linguistic information and has achieved significant success in various fields~\cite{wang2024t, ye2024altdiffusion, long2024generative}. However, the basis of these models is usually a known Transformer network, resulting in a square level computational complexity~\cite{jiang2024delving}. In order to improve the efficiency of the base model, ML-Mamba is proposed, which is an MLLM with linear computational complexity. Specifically, ML-Mamba integrates the efficient Mamba-2 language model into visual modalities and explores different modal fusion strategies to create effective multimodal Mamba-2~\cite{Dao2024TransformersAS}. Experiments have shown that ML-Mamba not only competes with current computationally efficient MLLMs such as LLaVA Phi, TinyLaVA, and MobileVLM v2, but also runs faster due to its linear sequence modeling characteristics. Interestingly, the results of the closed set prediction benchmark test show that ML-Mamba performs well in overcoming visual illusions and spatial relationship judgments, even comparable to LLaVA in performance with only 40\% of its parameters.

In terms of MLLMs for instruction tuning, recent research~\cite{karamcheti2024prismatic} has questioned the necessity of the pre alignment phase in MLLM training, pointing out that directly fine-tuning the entire LLM backbone and projector may be sufficient. In line with this, ML-Mamba only underwent a small amount of alignment training and then fine-tuned it on a large combination dataset containing visual multi-turn dialogues and visual alignment instructions.

\subsection{Mamba in the field of vision}
The successful application of Mamba in natural language processing (NLP) has inspired its adoption in visual applications~\cite{ren2024autoregressivepretrainingmambavision}. Vision Mamba (Vim)~\cite{Zhu2024VisionME} utilizes Vim blocks composed of pure Mamba layers: each Vim block models bidirectional representations using forward and backward scanning, and alleviates direction sensitivity issues in Mamba. Another approach, VMamba~\cite{Liu2024VMambaVS} utilizes Visual State Space (VSS) blocks that integrate Mamba and 2D convolutional layers, supported by a pyramid architecture similar to Swin Transformer~\cite{swin}: each VSS block first models 2D local information through 2D deep convolution as a token mixer, and then processes 2D global information horizontally and vertically through a cross scan module. Mamba ND~\cite{li2024mamba} further extends the functionality of Mamba to multidimensional data including images and videos. LocalMamba~\cite{localmamba}  segments the input image into multiple local windows and executes a state space model (SSM) in various directions within these windows to enhance local processing capabilities. EfficientVMamba~\cite{pei2024efficientvmamba} introduced an efficient 2D scanning technique that reduces computational requirements by performing atrous sampling on feature map blocks. In addition to these newly designed Mamba architectures, our work also draws inspiration from VL-Mamba~\cite{qiao2024vlmambaexploringstatespace}, a multimodal large language model based on state space models, which has shown great potential for long-sequence modeling with fast inference and linear scaling in sequence length. Compared with these newly designed Mamba architectures, our architecture closely follows Mamba's design ideas in the field of vision, enhancing the extraction of visual features with the latest Mamba-2 module. Our main goal with the Mamba-2 based architecture is to enhance multimodal representation and inference capabilities.

\section{Method}
\label{sec:method}
In this section, we first introduce the basic concepts of State Space Models (SSMs) (Sec. \ref{subsec:pre}). Subsequently, we provide a detailed description of the proposed ML-Mamba method (Sec. \ref{subsec:model}), which mainly comprises a visual encoder, a multi-modal connector called the Mamba-2 Scan Connector (MSC), an MLP projector, and the Mamba-2 large language model.

\subsection{Mamba Preliminaries} 
\label{subsec:pre}
The Mamba architecture is derived from state space sequence models~\cite{ssm1}, which models a $1$-D function or sequence $x(t)\in \mathbb{R}\to y(t)\in \mathbb{R}$ at time $t$ via expanded hidden states $h_t\in \mathbb{R}^N$. These hidden states evolve over time according to parameters $\mathbf{A}, \mathbf{B}, \mathbf{C}$ and are governed by linear ordinary differential equations (ODEs):
\begin{equation}\label{eq:1}
    \begin{split}
        h^{\prime}(t) &= \textbf{A}h(t) + \textbf{B}x(t),\\
        y(t) &= \textbf{C}h(t).
    \end{split}
\end{equation}
To discretize parameters in this continuous system, a common approach is to introduce a time scale parameter $\mathbf{\Delta}$ to transform continuous $\textbf{A}, \textbf{B}$ into discrete $\overline{\textbf{A}}, \overline{\textbf{B}}$ using the zero-order hold (ZOH) model~\cite{oppenheim1997signals}:
\begin{equation}
    \begin{aligned}
        \overline{\textbf{A}} &= \exp (\mathbf{\Delta} \textbf{A}),\\
        \overline{\textbf{B}} &= (\mathbf{\Delta} \textbf{A})^{-1}(\exp (\mathbf{\Delta} \textbf{A}) - \textbf{I})\cdot \mathbf{\Delta}\textbf{B}.
    \end{aligned}
\end{equation}
Using this transformation, Eq.~\ref{eq:1} can be rewritten as:
\begin{equation}
    \begin{aligned}
        h^{\prime}(t) &= \overline{\textbf{A}}h_{t-1} + \overline{\textbf{B}}x_t,\\
        y_t &= \textbf{C}h_t.
    \end{aligned}
\end{equation}
We then utilize the matrix $\overline{\textbf{K}}$ to enable efficient computation:
\begin{equation}
    \begin{aligned}
        \overline{\textbf{K}} &= (\textbf{C}\overline{\textbf{B}}, \textbf{C}\overline{\textbf{AB}}, ..., \textbf{C}\overline{\textbf{A}}^k\overline{\textbf{B}},...),\\
        \textbf{y} &= \textbf{x}*\overline{\textbf{K}},
    \end{aligned}
    \label{eq:mamba}
\end{equation}
where $k\in [0, L)$ and $L$ is the input sequence length. We also have $\textbf{y} = \{y_1, ..., y_L\}$, $\textbf{x} = \{x_1, ..., x_L\}$, while $\overline{\textbf{K}}\in \mathbb{R}^L$ can be regarded as the convolutional kernel.

By combining the modified parallel Mamba blocks with using SSD as the inner SSM layer, the Mamba-2 architecture is formed (as shown in Fig.~\ref{fig:arch}(a)). The performance of Mamba-2 models of varying sizes on the Pile dataset shows that it matches or outperforms Mamba and other open-source Transformer models on standard downstream evaluations.

\begin{figure}[t]
  \centering
  \includegraphics[width=0.75\linewidth]{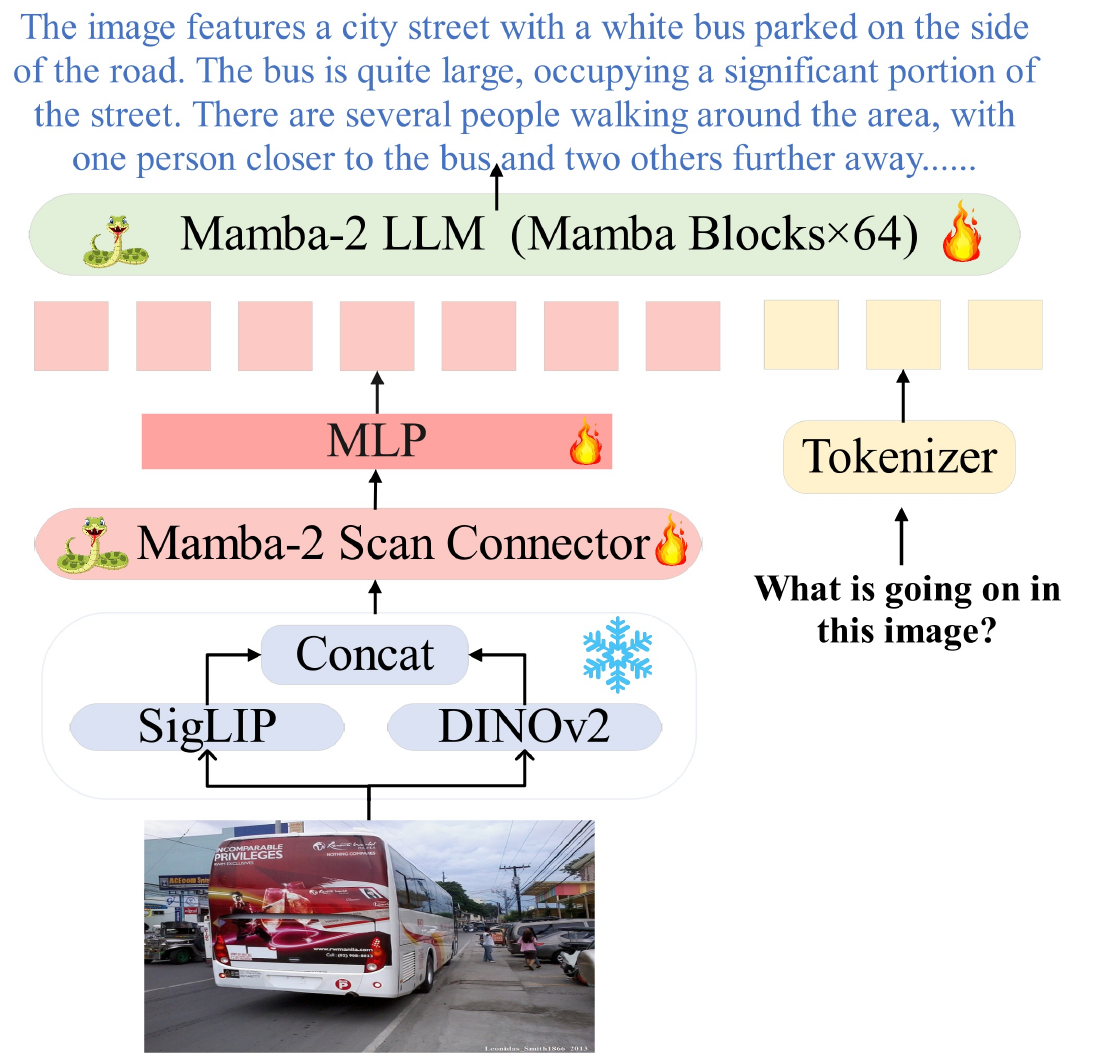}
  \caption{The architecture of ML-Mamba (right) uses Mamba-2 as the backbone (left). It includes a visual encoder, a multi-modal connector called the Mamba-2 Scan Connector (MSC), an MLP projector, and a language model. We use the pre-trained Mamba-2 large language model (Mamba-2 LLM) as the language model and a pre-trained visual transformer model as the visual encoder.
  }
  \label{fig:overall}
\end{figure}

\subsection{ML-Mamba Model}
\label{subsec:model}

\subsubsection{Overall Architecture}
\label{subsubsec:all}
The architecture of Mamba consists of four main components: a pre-trained visual encoder, a randomly initialized multi-modal connector called Mamba-2 Scan Connector (MSC), and a pre-trained large language model (Mamba-2 LLM), as shown in Fig.~\ref{fig:overall}. With an image as input, visual features are first extracted through the visual encoder. The extracted sequence of visual features is then fed into the multi-modal connector (MSC), whose output is mapped to the LLM using a multi-layer perceptron (MLP) projector. The output vector from the visual projector is then combined with tokenized text queries and input into the Mamba-2 LLM. Finally, the Mamba-2 LLM generates the corresponding response.

\subsubsection{Vision Encoder}

We integrate DINOv2~\cite{oquab2024dinov2} and SigLIP~\cite{zhai2023sigmoid} to serve as our vision backbone. The rationale behind this fusion is that combining the low-level spatial features captured by DINOv2 with the semantic features provided by SigLIP enhances performance on downstream tasks~\cite{tong2024eyes, karamcheti2024prismatic}. Given an input image $X_v \in \mathbb{R}^{C \times H \times W}$, the vision encoder divides the image into $N_v = HW/P^{2}$ patches of equal size, where $P^2$ represents the patch size. Both vision encoders process the patchified image as an input token sequence and concatenate their outputs to form compact visual representations $V_{img} \in \mathbb{R}^{N_v \times D_v}$:
  \begin{equation}
      V_{img} = [\varphi_{\rm SigLIP}(X_v); \varphi_{\rm DINOv2}(X_v)],
  \end{equation}
These outputs are then channeled to a dedicated task-specific head, with $D_v$ representing the dimensionality of the tokens generated as described above.

\begin{figure}[t]
  \centering
  \includegraphics[width=0.9\linewidth]{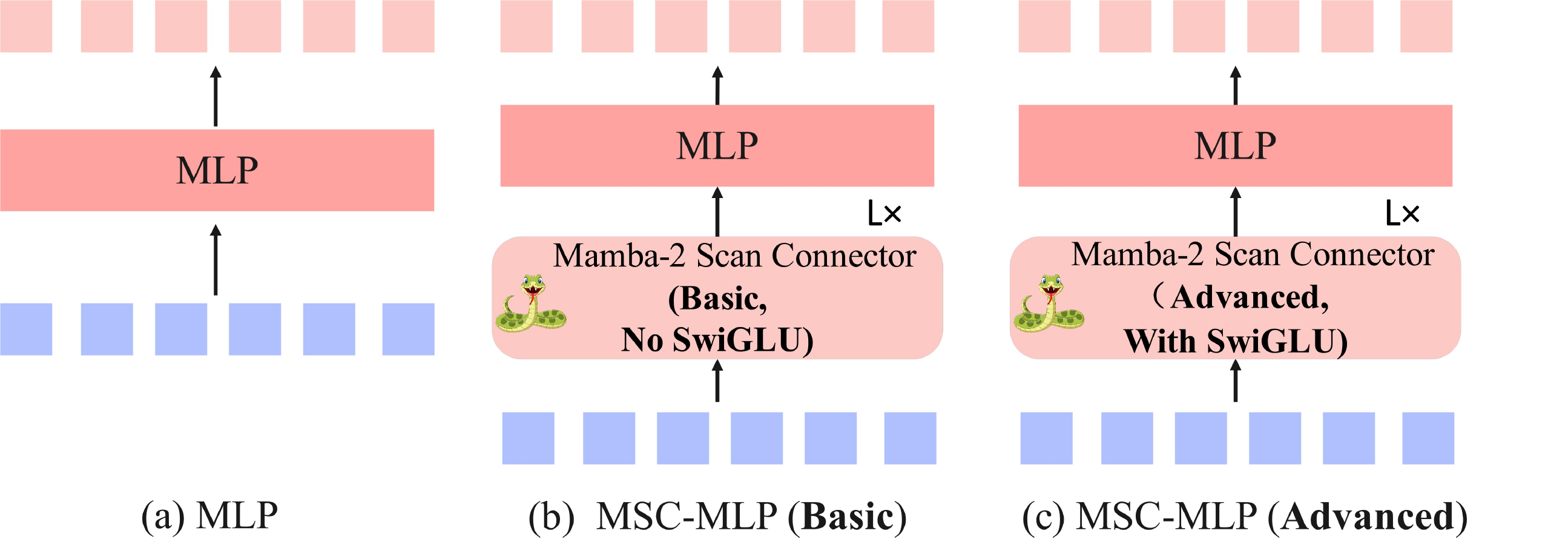}
  \caption{{Three architectures of MultiModal Connector: (a) MLP; (b) MSC-MLP (Basic); (c) MSC-MLP (Advanced). }
  }
  \label{fig:mc}
  \vspace{-10pt}
\end{figure}

\begin{figure}[t]
 \vspace{-10pt}
  \centering
  \includegraphics[width=0.8\linewidth]{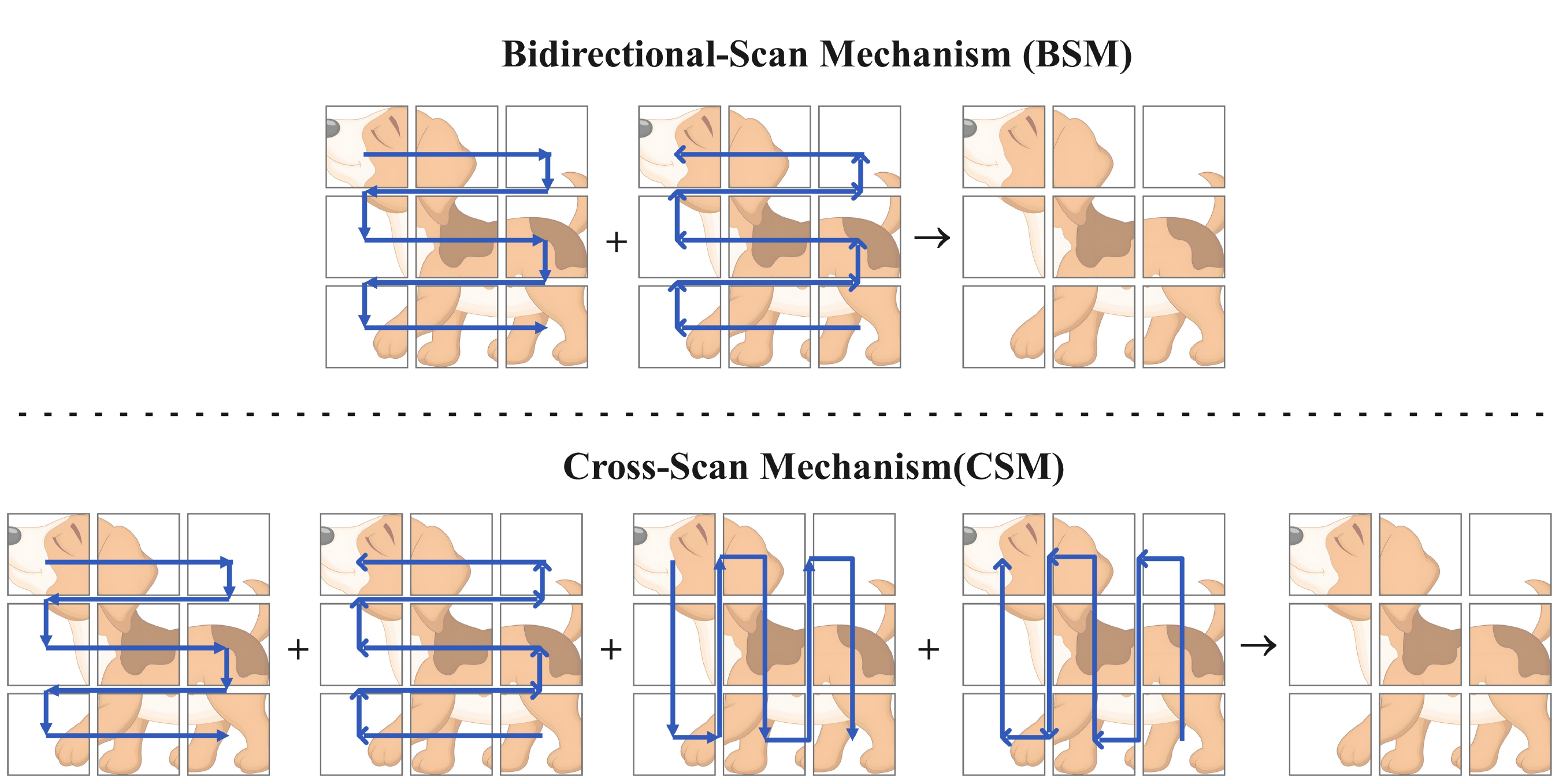}
  \caption{Illustration of two different Vision Selective Scan (VSS) Mechanisms: Bidirectional-Scan Mechanism (BSM) (top) and Cross-Scan Mechanism (CSM) (bottom).
  }
  \label{fig:2D scan}
\end{figure}

\subsubsection{MultiModal Connector}

Multimodal connectors act between visual features and language models to ensure seamless integration of visual and linguistic information. In this study, we explored a novel multimodal connector called Mamba-2 Scan Connector (MSC) architecture aimed at addressing the challenge of unclear causal relationships in computer vision. The traditional state space model (SSM) is typically used to process sequence data with causal relationships, such as language sequences, but this approach is clearly not applicable to non causal visual sequences generated by visual encoders.

The core of the MSC module is a combination of the two-dimensional Mamba-2 visual selective scanning (MVSS) module and the SwiGLU module. We attempted to integrate this module into the multimodal connector of the ML-Mamba multimodal learning framework.

\begin{figure}
    \centering
    \resizebox{4cm}{6cm}{\includegraphics{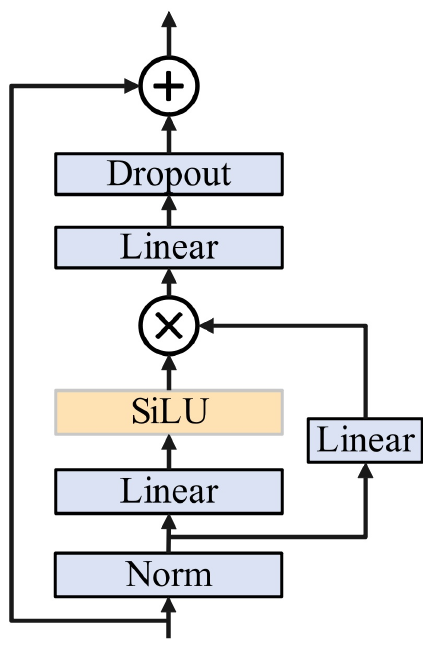}}
    \caption{SwiGLU structure in MSC-MLP (Advanced).}
    \label{fig:swiglu}
\end{figure}

Specifically, we studied three variants of multimodal connectors:

\begin{itemize}
    \item \textbf{MLP}: a three-layer Multi-Layer Perceptron (MLP) (see Fig.~\ref{fig:mc}(a)) that aligns the features of vision and text.
    \item \textbf{MSC-MLP (Basic)}: It combines the multimodal connector called the Mamba-2 Scan Connector (MSC) module, which does not include the SwiGLU module and is intended to enhance the processing capability of two-dimensional non-causal visual information. Subsequently, the MLP aligns the features of vision and text (see Fig.~\ref{fig:mc}(b))).
    \item \textbf{MSC-MLP (Advanced)}: This variant combines the MSC module and MLP, where the MSC module includes the SwiGLU (see Fig.~\ref{fig:swiglu}) module for more complex feature extraction and pattern learning (see Fig.~\ref{fig:mc}(c)). 
\end{itemize}

\begin{figure}
    \centering
    \includegraphics[width=\linewidth]{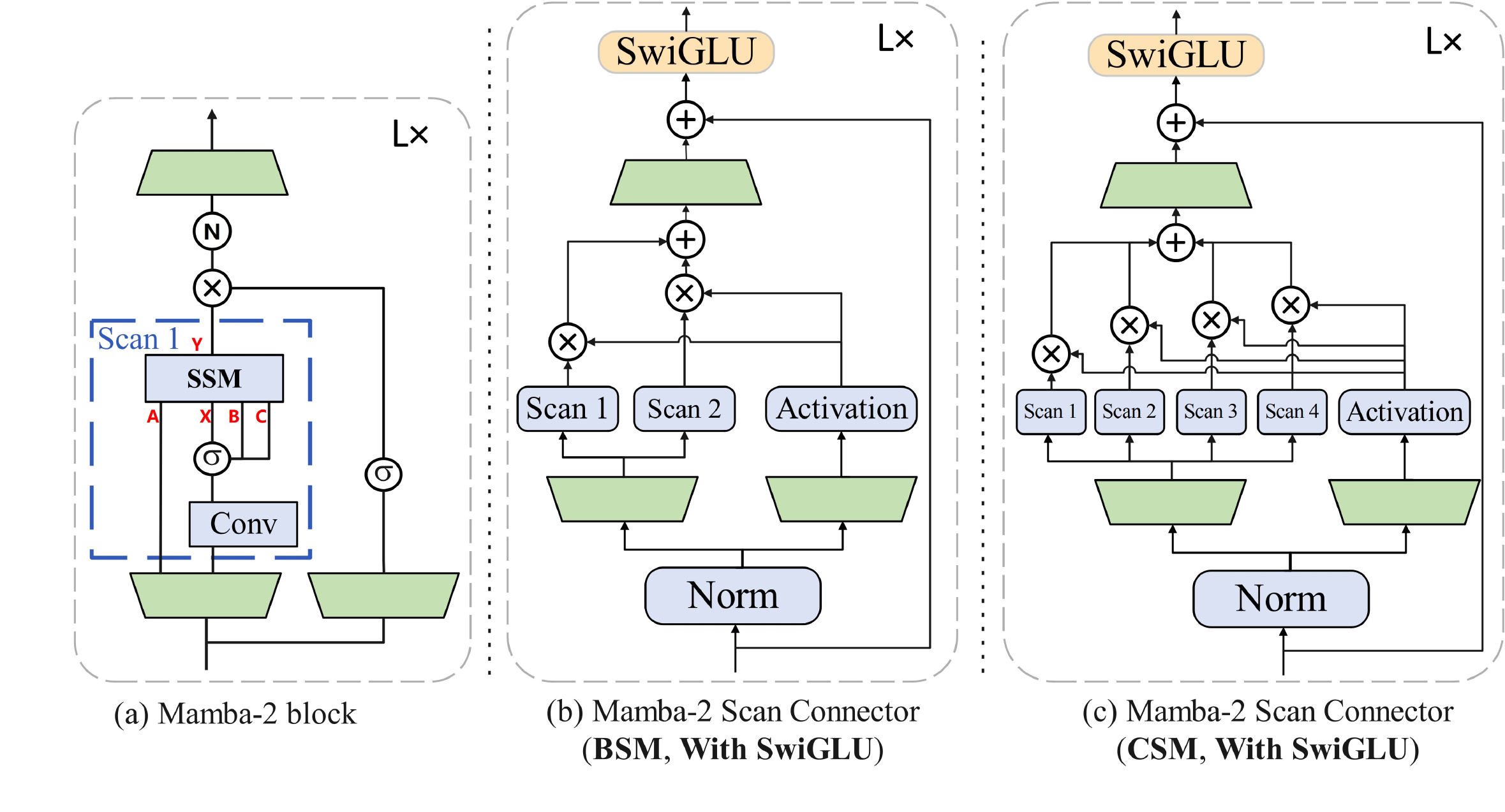}
    \vspace{-5mm}
    \caption{The comparison of block architectures between Mamba-2 block, and Mamba-2 Scan Connector
(BSM, With SwiGLU) and Mamba-2 Scan Connector (CSM, With SwiGLU).}
    \label{fig:arch}
\end{figure}

The MSC module bridges the gap between 1D sequential processing capability (typical of SSM) and 2D non causal visual information by introducing two 2D scanning mechanisms. These scanning mechanisms include:

\begin{itemize}
    \item \textbf{Bidirectional-Scan Mechanism (BSM)}: Scanning the complementary features of the image in both forward and backward directions to capture a broader context without increasing computational complexity (shown at the top of Fig.~\ref{fig:2D scan}).The corresponding model structure is depicted in Fig.~\ref{fig:arch}(b).
    \item \textbf{Cross-Scan Mechanism (CSM)}: unfolds image patch features into sequences along rows and columns and scans them in four directions (diagonally across the image) (shown at the bottom of Fig.~\ref{fig:2D scan}). The corresponding model structure is depicted in Fig.~\ref{fig:arch}(c).
\end{itemize}

After scanning, these feature sequences are processed by the Mamba-2 layer and reshaped into the patch order of the original image, and finally merged into a comprehensive representation for subsequent multimodal learning tasks. The goal of this method is to improve the modeling ability of complex visual data, especially when it involves multimodal input and nonlinear relationship modeling, to enhance the performance and robustness of computer vision tasks.

As shown in Fig.~\ref{fig:mc}(a), the input of the multimodal connector is the sequential image patch features $V_{img}$ extracted from the input images via the transformer-based vision encoder. These feature vectors are then passed into a three-layer Mult-Layer (MLP): 
\begin{equation}
\begin{aligned}
\label{eq:mmc}
V_{out} &= \mathbf{MLP}(V_{img}).
\end{aligned}
\end{equation}
As shown in Fig.~\ref{fig:mc}(b), the input of the multimodal connector is the sequential image patch features $V_{img}$ extracted from the input images via the transformer-based vision encoder. These feature vectors are then passed through a Mamba-2 Scan Connector (MSC) module to obtain the visual scanned feature $V_{scan}$.
After the MSC module, the output vectors $V_{scan}$ is then passed into a three-layer Mult-Layer (MLP): 
\begin{equation}
\begin{aligned}
\label{eq:mmc}
V_{scan} &= \mathbf{MSC_{Basic}(V_{img})}, \\
V_{out} &= \mathbf{MLP}(V_{scan}).
\end{aligned}
\end{equation}
As shown in Fig.~\ref{fig:mc}(c), the feed-forward pass progress can be formulated as follows:
\begin{equation}
\begin{aligned}
\label{eq:mmc}
V_{scan} &= \mathbf{MSC_{Basic}(V_{img})}, \\
V_{scan}^{'} &= \mathbf{SwiGLU(V_{scan})}, \\
V_{out} &= \mathbf{MLP}(V_{scan}^{'}).\\
\end{aligned}
\end{equation}

\subsubsection{{Mamba-2 Large Language Model}}

The Mamba-2 language model~\cite{Dao2024TransformersAS} serves as the primary language processing component responsible for understanding and generating text. The workflow design of the visual encoder and multimodal connector ensures that visual information can be effectively transmitted to the Mamba-2 language model, enabling the model to process and understand complex multimodal data.
\begin{equation}
\begin{aligned}
\label{eq:llm}
R = {f_{L}}(V_{out}, f_T(Q)).
\end{aligned}
\end{equation}

\subsubsection{Training Process}

We first use a 558K subset of the LAION-CC-SBU dataset to align the Mamba-2 Scan Connector (MSC) and the MLP projector. During the fine-tuning stage, we simultaneously optimized the Mamba-2 Scan Connector (MSC), the projector, and the Mamba LLM. This comprehensive training effort was executed on 8 NVIDIA A100 GPUs. The fine-tuning was conducted over two epochs, randomly sampling from the Mixed Dataset Used in LLaVA v1.5, which includes a total of 665K visual multi-round dialogue samples and pure text dialogue data.
\section{Experiment}
\label{sec:expri}
We conducted a comprehensive experimental evaluation of ML-Mamba through four aspects: benchmarking evaluation: We used six commonly used visual language model (VLM) benchmarks to evaluate the effectiveness of the proposed method. These benchmarks include four open-ended visual question answering tasks that require different reasoning abilities, as well as two closed set prediction tasks that involve determining spatial relationships of objects and detecting visual illusions.

\begin{itemize}
    \item \textbf{Efficiency evaluation}: We conducted a comparative evaluation of ML-Mamba and other Transformer based models at similar model sizes to validate our model's improvement in efficiency.
    \item \textbf{Ablation study}: We further explored some design choices in the model structure through ablation studies to determine which components have a significant impact on model performance.
    \item \textbf{Comparison of answer generation quality}: We have provided specific examples to demonstrate the comparison of our model with other models in terms of answer generation quality. Through these experiments, we comprehensively evaluated the performance and advantages of ML-Mamba. 
\end{itemize}

\subsection{Experimental Setup}
\label{subsec:setup}

Table~\ref{tab:setting} details the hyperparameters of the ML-Mamba model. For the visual encoder part, DINOv2 adopts the same ViT structure as in its original paper, namely a ViT-Large model with 304M parameters, pretrained on the LVD-142M dataset. SigLIP uses a slightly larger shape-optimized version than ViT-Large. The resolution of the input images is set to 384x384, with the number of visual tokens being 729.

The backbone of the LLM is initialized using the pretrained weights from the Mamba-2 model, while the multimodal connectors (MSC) and projectors are always randomly initialized. We chose an open-source model weight from the Huggingface platform to initialize our model as the LLM backbone for our proposed model.

The entire training process took approximately 31 hours on 8 NVIDIA A100 80GB GPUs. During training, we used Pytorch's fully shared data parallel framework~\cite{zhao2023fsdp} and adopted automatic mixed precision with FP32 and BF16 for distributed training. The batch size was set to 64. We used the AdamW~\cite{loshchilov2019adamw} optimizer and updated the network parameters using a learning rate with cosine decay. The learning rate was set to $2\times10^{-5}$, the decay factor was 0.1, and the warm-up ratio was 0.03. The model was trained for 2 epochs with supervised fine-tuning.

\begin{table}[th!]
  \caption{The configuration of the model and hyperparameters for training.}
  \label{tab:setting}
  \centering
  \renewcommand{\arraystretch}{0.9}
  \setlength{\tabcolsep}{4pt}
  \begin{tabular}{@{}l@{\hspace{2em}}l@{}}
    \toprule
    Configuration \\
    \midrule
    Vision Encoder & DINOv2 + SigLIP\\
    LLM init & Mamba-2 2.7b\\
    MLP + Mamba-2 Scan Connector init & Random\\
    Image resolution & $384 \times 384$\\
    Alignment / Fine-Tuning Samples & 558K / 665K\\
    Optimizer & AdamW\\
    LR schedule & Cosine decay\\
    Learning Rate & 2e-5\\
    Weight decay & 0.1\\
    Warmup ratio & 0.03\\
    Alignment and Fine-tuning epochs & 1 each\\
  \bottomrule
  \end{tabular}
\end{table}

\begin{table*}[th!]
\caption{\textbf{Comparison with SoTA methods on six benchmarks:} 
VQA-v2~\cite{goyal2017vqav2}; GQA~\cite{hudson2019gqa}; VQA$^\text{T}$: TextVQA~\cite{singh2019textvqa}; POPE~\cite{li2023pope}; VizWiz~\cite{gurari2018vizwiz}; VSR~\cite{liu2023visual}.
PT and IT indicate the number of samples in the pretraining and instruction tuning stages, respectively.
}

\label{tab:results}
\centering
\renewcommand{\arraystretch}{1}
\setlength{\tabcolsep}{3pt} 
\normalsize 
\begin{tabular}{ll cc | cccccc }
\toprule
Method & LLM  & PT & IT & VQA$^\text{v2}$ & GQA & VQA$^\text{T}$ & POPE & VizWiz & VSR \\
\midrule
\textcolor{gray}{BLIP-2~\cite{Li2023BLIP2BL}} & \textcolor{gray}{Vicuna-13B}  & \textcolor{gray}{129M} & \textcolor{gray}{-} & \textcolor{gray}{41.0} & \textcolor{gray}{41.0} & \textcolor{gray}{42.5} & \textcolor{gray}{85.3} & \textcolor{gray}{19.6} & \textcolor{gray}{50.9} \\
\textcolor{gray}{MiniGPT-4~\cite{zhu2023minigpt}} & \textcolor{gray}{Vicuna-7B} & \textcolor{gray}{5M} & \textcolor{gray}{5K} & \textcolor{gray}{32.2} & \textcolor{gray}{32.2} & \textcolor{gray}{-} & \textcolor{gray}{-} & \textcolor{gray}{-} & \textcolor{gray}{-} \\
\textcolor{gray}{InstructBLIP~\cite{instructblip}} & \textcolor{gray}{Vicuna-7B} & \textcolor{gray}{129M} & \textcolor{gray}{1.2M} & \textcolor{gray}{--} & \textcolor{gray}{49.2} & \textcolor{gray}{50.1} & \textcolor{gray}{--} & \textcolor{gray}{34.5} & \textcolor{gray}{54.3} \\ 
\textcolor{gray}{InstructBLIP~\cite{instructblip}} & \textcolor{gray}{Vicuna-13B}  & \textcolor{gray}{129M} & \textcolor{gray}{1.2M} & \textcolor{gray}{--} & \textcolor{gray}{49.5} & \textcolor{gray}{50.7} & \textcolor{gray}{78.9} & \textcolor{gray}{33.4} & \textcolor{gray}{52.1} \\ 
\textcolor{gray}{Shikra~\cite{chen2023shikra}} & \textcolor{gray}{Vicuna-13B}  & \textcolor{gray}{600K} & \textcolor{gray}{5.5M} & \textcolor{gray}{77.4} & \textcolor{gray}{--} & \textcolor{gray}{--} & \textcolor{gray}{--} & \textcolor{gray}{--} & \textcolor{gray}{--} \\ 
\textcolor{gray}{IDEFICS-9B~\cite{laurenccon2024obelics}} & \textcolor{gray}{LLaMA-7B} & \textcolor{gray}{353M} & \textcolor{gray}{1M} & \textcolor{gray}{50.9} & \textcolor{gray}{38.4} & \textcolor{gray}{25.9} & \textcolor{gray}{--} & \textcolor{gray}{35.5} & \textcolor{gray}{-} \\ 
\textcolor{gray}{IDEFICS-80B~\cite{laurenccon2024obelics}} & \textcolor{gray}{LLaMA-65B} & \textcolor{gray}{353M} & \textcolor{gray}{1M} & \textcolor{gray}{60.0} & \textcolor{gray}{45.2} & \textcolor{gray}{30.9} & \textcolor{gray}{--} & \textcolor{gray}{36.0} & \textcolor{gray}{-} \\
\textcolor{gray}{Qwen-VL~\cite{bai2023qwen}} & \textcolor{gray}{Qwen-7B}  & \textcolor{gray}{1.4B} & \textcolor{gray}{50M} & \textcolor{gray}{78.8} & \textcolor{gray}{59.3} & \textcolor{gray}{63.8} & \textcolor{gray}{--} & \textcolor{gray}{35.2} & \textcolor{gray}{-} \\ 
\textcolor{gray}{Qwen-VL-Chat~\cite{bai2023qwen}}  & \textcolor{gray}{Qwen-7B} & \textcolor{gray}{1.4B} & \textcolor{gray}{50M} & \textcolor{gray}{78.2} & \textcolor{gray}{57.5} & \textcolor{gray}{61.5} & \textcolor{gray}{--} & \textcolor{gray}{--} & \textcolor{gray}{--} \\ 
\textcolor{gray}{LLaVA-1.5~\cite{liu2023llava}} & \textcolor{gray}{Vicuna-7B} & \textcolor{gray}{558K} & \textcolor{gray}{665K} & \textcolor{gray}{78.5} & \textcolor{gray}{62.0} & \textcolor{gray}{58.2} & \textcolor{gray}{85.9} & \textcolor{gray}{50.0} & \textcolor{gray}{-} \\ 
\textcolor{gray}{LLaVA-1.5~\cite{liu2023llava}} & \textcolor{gray}{Vicuna-13B} & \textcolor{gray}{558K} & \textcolor{gray}{665K} & \textcolor{gray}{80.0} & \textcolor{gray}{63.3} & \textcolor{gray}{61.3} & \textcolor{gray}{85.9} & \textcolor{gray}{-} & \textcolor{gray}{-} \\
\textcolor{gray}{TinyLLaVA~\cite{zhou2024tinyllava}} & \textcolor{gray}{Phi2-2.7B} & \textcolor{gray}{1804K} & \textcolor{gray}{1330K} & \textcolor{gray}{79.9} & \textcolor{gray}{62.0} & \textcolor{gray}{-} & \textcolor{gray}{86.4} & \textcolor{gray}{-} & \textcolor{gray}{-} \\
\midrule
LLaVA-Phi~\cite{zhu2024llavaphi} & Phi-2-2.7B & 558K& 665K& 71.4 & - & 48.6 & 85.0 & 35.9 & -\\
MobileVLM-3B~\cite{Chu2023MobileVLMA} & MobileLLaMA-2.7B  & 558K & 665K& - & {59.0} & 47.5 & 84.9 & - & -\\
Cobra~\cite{zhao2024cobraextendingmambamultimodal}& Mamba LLM{-2.8B} &558K & 665K&{75.19} & 58.7 & - & {87.2} &- &-\\
VL-Mamba~\cite{qiao2024vlmambaexploringstatespace}& Mamba LLM{-2.8B} &558K & 665K&{76.6} & 56.2 & {48.9} &{84.4}&-&-\\
\midrule
\textbf{ML-Mamba (ours)}& Mamba-2 LLM{-2.7B} &558K & 665K&{75.26} & \textbf{60.68} & \textbf{52.2} & \textbf{88.3} & \textbf{45.17} & \textbf{51.5}\\
\bottomrule
\end{tabular}

\end{table*}

\begin{figure}[h]
  \centering
  \includegraphics[width=0.99\linewidth]{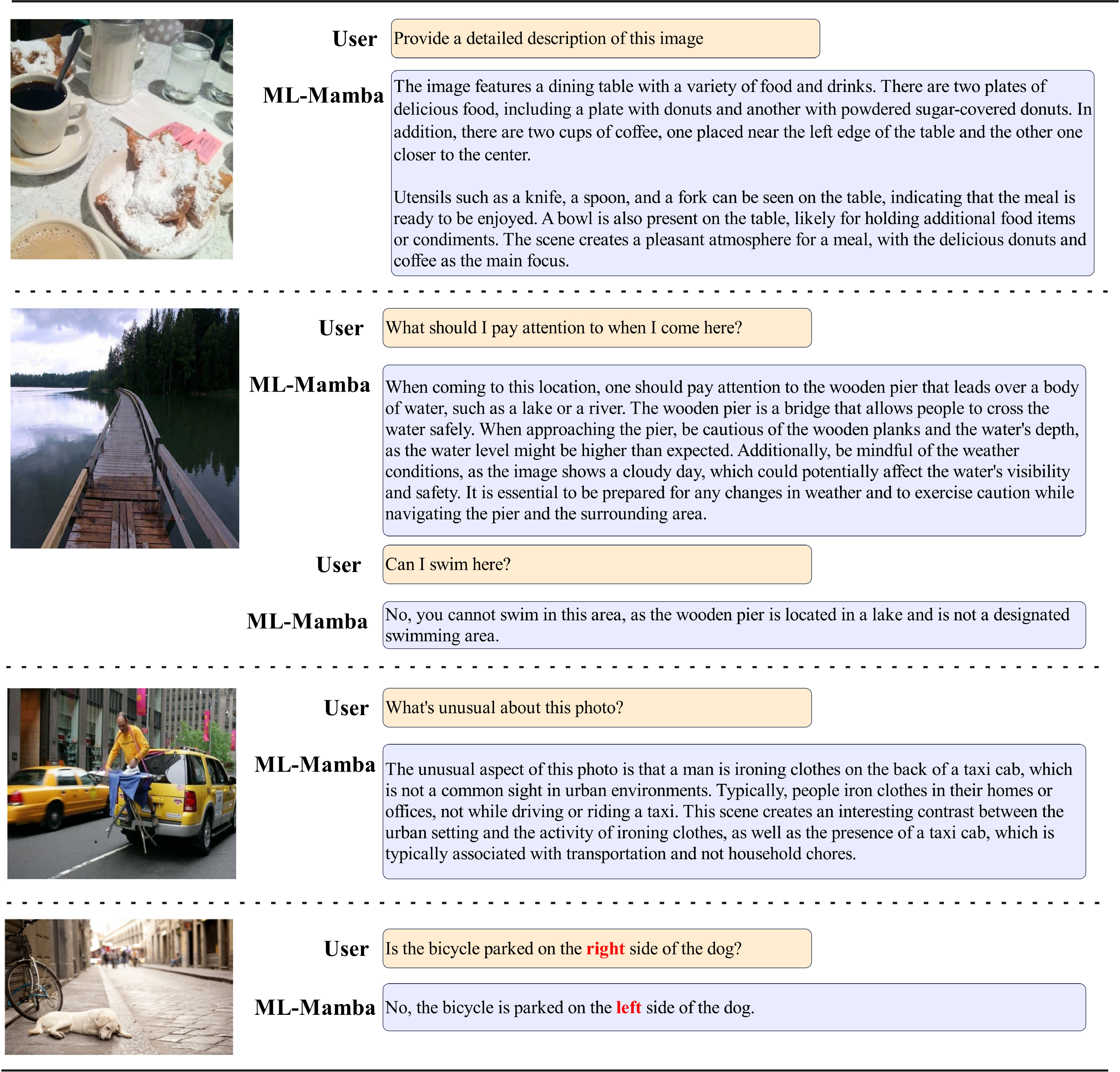}
  \caption{Examples of response generated by ML-Mamba.
  }
  \label{fig:chat}
\end{figure}

\subsection{Results}
\label{subsec:sota}

In addition, we further evaluated the model on six carefully designed metrics, particularly VizWiz~\cite{gurari2018vizwiz} and VQAv2~\cite{goyal2017making}, for assessing general visual reasoning ability. VizWiz includes common sense questions and unanswerable questions, requiring the model to avoid incorrect answers to evaluate its reliability. GQA evaluates spatial understanding and multi-step reasoning in real-world images. The issues in TextVQA are related to the text in the image, evaluating the model's optical character recognition (OCR) and inference capabilities. POPE provides a benchmark for evaluating object hallucinations and is a binary classification task that prompts the model to answer whether the object exists. We also introduced two closed set prediction benchmarks consisting of VSR~\cite{liu2023visual} and POPE~\cite{li2023evaluating}. VSR evaluates the model's ability to understand spatial relationships between different images, while POPE evaluates the VLM's ability to avoid severe illusion problems. VSR and POPE calculate scores based on the probability of providing the correct answer.

We evaluated VizWiz, VQAv2, and TextVQA using validation sets, while using the recommended test dev partition for GQA, zero sample test partition for VSR, and evaluation partition for POPE.

To demonstrate the model's effectiveness, we compared it with a VLM of the same scale with approximately 3B parameters, or with a larger VLM containing twice the number of parameters. As shown in Table~\ref{tab:results}, despite having only 40\% of LLaVA v1.5 7B's parameters, ML-Mamba performs comparably across multiple benchmarks and even surpasses all models on POPE.

As shown in Table~\ref{tab:results}, compared with VLM with similar parameter numbers, ML-Mamba consistently achieved better performance than LLaVA Phi in VQAv2, GQA, VQA$^\text{T}$, POPE and VizWiz. While VL-Mamba performs better on VQAv2, our ML-Mamba outperforms VL-Mamba on GQA, VQA$^\text{T}$, and POPE. MobileVLM is another parallel work aimed at producing small-scale LLMs, and is therefore also introduced in experiments. In summary, these results indicate that ML-Mamba matches the performance of state-of-the-art models at the same level ($\sim$3B) on multiple benchmarks and remains competitive when compared to larger scale models (7B and above).

We present some examples to illustrate the qualitative results of ML-Mamba. As shown in Fig.~\ref{fig:chat}, ML-Mamba effectively understands the user's questions and responds accurately.

\subsection{Reasoning speed}
In order to evaluate the efficiency advantage of the ML-Mamba model, especially the speed improvement brought by its linear sequence modeling, we conducted a detailed inference speed comparison experiment. In the experiment, we compared ML-Mamba with two baseline models of the same scale parameters, TinyLaVA 3B and MobileVLM v2 3B.

All models were evaluated in the same hardware environment, namely a single Nvidia A100 PCIe 80GB GPU. Each model receives the same example image as input, with a unified image resolution of 336 $\times$ 336 pixels, and is processed by a CLIP encoder. For TinyLaVA, the model receives 576 image markers processed by the projector; MobileVLM v2 reduces the number of image labels to 144 through LDP blocks. In contrast, ML-Mamba uses dual encoders to process images with a resolution of 384 $\times$ 384, resulting in an increase in the actual number of image labels processed to 729.

In the experiment, all models received the same question: "Provide a detailed description of the image." and set the number of output labels to 256. The total time is the entire process from image encoding until the complete answer is generated.And we calculated the average number of tokens generated per second by $Eval_{avg} = 256/ T_{total}$.

\begin{table*}[t]
\caption{Latency comparison of small-scale VLMs with $\sim$3B parameters.}
\label{tab:speed}
\centering
\renewcommand{\arraystretch}{1.2} 
\normalsize 
\begin{tabular}{l| cccc }
\toprule
Model & LM & {$Eval_{avg}$(tokens/s)} &  $Total$ (s) \\
\midrule
 TinyLLaVA & Phi-2 2.7B & 38 & 6.45 \\
 MobileVLM v2 & MobileLLaMA 2.7B & 50 & 5.15  \\
 ML-Mamba & Mamba-2 2.7B & \textbf{171} & \textbf{1.47} \\
\bottomrule
\end{tabular}
\end{table*}

The results from Table~\ref{tab:speed}  demonstrated that although the number of image markers processed by ML-Mamba significantly increased, it still exhibited extremely fast inference speed. Compared to MobileVLM v2, although the latter has undergone multiple lightweight optimizations, the time required for ML-Mamba to complete inference is only about 30\% of the former. This indicates that ML-Mamba not only maintains high speed while processing larger data, but also, thanks to the characteristics of its RNN like model, its memory usage does not significantly increase with the increase of image marker length, as such models maintain a fixed size hidden state to store historical information during the inference process.

The excellent performance of the ML-Mamba model in inference speed proves its advantage in linear sequence modeling, especially when dealing with a large number of image labels. Compared to Transformer based models, ML-Mamba demonstrates significant speed improvements, providing strong support for multimodal tasks that require rapid response.

\subsection{Ablation Study}
\label{subsec:abla}

\subsubsection{Effects of Language Model Variants}

Table~\ref{tab:llm} presents the results of ablation experiments evaluating the effectiveness of different language model variants. We conducted experiments on three different variants, namely Mamba-2 with parameters of 780m, 1.3b, and 2.7b, trained on the Pile dataset (containing 300B tokens). Specifically, we constructed a baseline model using the same variant of DINOv2+SigLIP as the visual encoder, Mamba-2 language model as the backbone of a large language model, and a regular MLP multimodal connector without a 2D visual selection scanning module. We can see that as the model size and number of training tokens increase, Mamba2-2.7B outperforms other variants on all benchmarks. Therefore, we chose Mamba2-2.7B for other experiments.

\begin{table*}[th!]
\caption{Ablation study of the variants of the language model.}
\label{tab:llm}
\centering
\renewcommand{\arraystretch}{1.2}
\normalsize
\begin{tabular}{l| cccccc }
\toprule
Method & VQA$^\text{v2}$ & GQA & VQA$^\text{T}$ & POPE & VizWiz & VSR \\
\midrule
 Mamba2 - 780m & 71.7    & 51.92  & 48.1   & 81.6   & 41.5  & 47.7 \\
 Mamba2 - 1.3b & 73.6    & 55.41  & 50.8   & 83.7   & 43.7  & 49.3 \\
 Mamba2 - 2.7b & {75.26} & 60.68 & {52.2} & {88.3} & 45.17 & 51.5 \\
\bottomrule
\end{tabular}
\end{table*}

\subsubsection{Effects of Different Visual Encoders}
Recent research has found that although language image models similar to CLIP can provide rich semantic information, they may lose detailed information about the image itself. Therefore, we further introduce DINOv2 as a supplementary encoder and connect the visual representations of these two encoders for subsequent LLM. As shown in Table~\ref{tab:visenc}, the introduction of DINOv2 significantly improved the model performance in six benchmark tests. This result suggests a meaningful principle when selecting a visual encoder for downstream tasks. Therefore, we ultimately chose DINOv2+SigLIP as the visual encoder to construct our model and used it for further experiments. Through this combination, we can achieve better performance on multiple benchmarks.

\begin{table*}[th!]
\caption{Ablation study of the vision encoder.
}
\label{tab:visenc}
\centering
\renewcommand{\arraystretch}{1.15}
\resizebox{0.85\linewidth}{!}{
\begin{tabular}{l| cccccc }
\toprule
Method & VQA$^\text{v2}$ & GQA & VQA$^\text{T}$ & POPE & VizWiz & VSR \\
\midrule
DINOv2          & 73.73    & 58.84  & 51.13   & 86.6   & 44.23 & 50.73 \\
SigLIP          & 74.61    & 59.43  & 50.78   & 87.4   & 45.07 & 50.54 \\
DINOv2 + SigLIP & 75.26    & 60.68  & 52.20   & 88.3   & 45.17 & 51.50 \\
\bottomrule
\end{tabular}
}
\end{table*}

\subsubsection{Ablation on different multimodal connector structures}
We also explored the impact of different architectures of multi-mode connectors. We evaluated three different MMC variants: MLP, MSC-MLP (Basic), and MSC-MLP (Advanced).
As shown in Table ~\ref{tab:arch-mmc}, by comparing these three architectures, we observed that MSC-MLP (Advanced) performed relatively better on most benchmark tests, especially on VQA, demonstrating the effectiveness of combining MSC modules with swiGLU. Note that these models use DINOv2+SigLIP as the visual encoder, Mamba2-2.7B as the language model, and a bidirectional selective scanning mechanism. Consequently, we ultimately chose MSC-MLP (Advanced) as our model and used it for further experiments.

\begin{table*}[th!]
\caption{Ablation study of the different architectures of  multimodal connector.
}
\label{tab:arch-mmc}
\centering
\renewcommand{\arraystretch}{1.15}
\resizebox{0.85\linewidth}{!}{
\begin{tabular}{l| cccccc }
\toprule
Method & VQA$^\text{v2}$ & GQA & VQA$^\text{T}$ & POPE & VizWiz & VSR \\
\midrule
 MLP               & 73.42    & 58.87  & 50.31   & 86.1   & 43.87 & 50.13 \\
 MSC-MLP(Basic)    & 75.09    & 60.14  & 51.72   & 86.5   & 44.57 & 50.76 \\
 MSC-MLP(Advanced) & 75.26    & 60.68  & 52.20   & 88.3   & 45.17 & 51.50 \\
\bottomrule
\end{tabular}
}
\end{table*}

\subsubsection{Under different scanning mechanisms}
We compared the bidirectional scanning mechanism (BSM) and cross scanning mechanism (CSM) in MMC modules. As shown in Table ~\ref{tab:scan}, although BSM and CSM perform similarly in some benchmark tests, such as scoring 76.6 in one test, BSM shows superior performance in most benchmark tests. This highlights its advantages in handling 2D visual information for multimodal learning tasks.

\begin{table*}[th!]
\caption{Ablation study of the scan mechanisms.
}
\label{tab:scan}
\centering
\renewcommand{\arraystretch}{1.15}
\resizebox{0.99\linewidth}{!}{
\begin{tabular}{l| cccccc }
\toprule
Method & VQA$^\text{v2}$ & GQA & VQA$^\text{T}$ & POPE & VizWiz & VSR \\
\midrule
 Bidirectional-Scan Mechanism (BSM) & 75.26    & 60.68  & 52.20   & 88.3   & 45.17 & 51.50 \\
 Cross-Scan Mechanism (CSM)         & 75.14    & 60.13  & 52.31   & 88.5   & 44.89 & 51.14 \\
\bottomrule
\end{tabular}
}
\end{table*}

\section{Limitation}
The training of ML-Mamba relies on specific multimodal datasets, which may have biases or incomplete coverage in certain aspects. Developing more comprehensive and diverse datasets, as well as improving data preprocessing and augmentation techniques, will help enhance the generalization ability and applicability of ML-Mamba in different scenarios.

ML-Mamba currently faces challenges in running on mobile devices, especially in meeting the memory usage requirements of these devices. In order to make ML-Mamba run more smoothly on devices such as smartphones or tablets, further optimization, especially for low memory environments, is necessary.

\section{Conclusion}
This article introduces a novel multimodal learning model, ML-Mamba, which utilizes the latest state space model (SSM) Mamba-2 to solve multimodal learning tasks. It uses a pre-trained Mamba-2 language model as the language model and introduces the multimodal connector Mamba-2 Scan Connector (MSC) module to bridge the gap between 2D non-causal image information and the inherent causal modeling ability of SSM. By conducting comprehensive experiments and ablation studies, ML-Mamba performed well in multimodal benchmark testing, demonstrating its effectiveness and the potential of SSM in multimodal learning. On the other hand, ML-Mamba addresses the efficiency bottleneck of existing multimodal large language models by using models with linear computational complexity. This significantly improves computational efficiency and excels in visual illusion and spatial relationship judgment while reducing the number of parameters. These advancements open new possibilities for deploying high-performance AI models in environments that process visual information at high frequencies.

\bibliographystyle{abbrv}
{
	\small
	\bibliography{main}
}

\end{document}